\documentclass{bmvc2k}
\usepackage{amsmath}
\usepackage{graphicx}
\usepackage{color} \usepackage{colortbl}
\definecolor{intnull2}{RGB}{214,214,214} 
\usepackage{amssymb}
\usepackage{subfigure}
\usepackage{mathtools}

\def\etal{\emph{et al}\bmvaOneDot}

\title{SP-NET: One Shot Fingerprint Singular-Point Detector}

\addauthor{Geetika Arora}{p2016406@pilani.bits-pilani.ac.in}{1}
\addauthor{Ranjeet Ranjan Jha}{ranjanjharanjeet@gmail.com}{2}
\addauthor{Akash Agrawal}{akashagrawal01010@gmail.com}{2}
\addauthor{Kamlesh Tiwari}{kamlesh.tiwari@pilani.bits-pilani.ac.in}{1}
\addauthor{Aditya Nigam}{aditya@iitmandi.ac.in}{2}

\addinstitution{Department of CSIS, Birla Institute of Technology and Science Pilani, Pilani-333031, Jhunjhunu, Rajasthan, INDIA}
\addinstitution{School of Computing and Electrical Engineering, Indian Institute of Technology Mandi, Mandi-175005, Himachal Pradesh, INDIA}

\runninghead{G Arora, RR Jha \etal}{SP-NET: Fingerprint Singular-Point Detector}

\def\BibTeX{{\rm B\kern-.05em{\sc i\kern-.025em b}\kern-.08em
    T\kern-.1667em\lower.7ex\hbox{E}\kern-.125emX}}

\begin{document}
	
\maketitle 

\begin{abstract}
     Singular points of a fingerprint image are special locations having high curvature properties. They can play a pivotal role in fingerprint normalization and reliable feature extraction. Accurate and efficient extraction of a singular point plays a major role in successful fingerprint recognition and indexing. In this paper, a novel deep learning based architecture is proposed for one shot (end-to-end) singular point detection from an input fingerprint image. The model consists of a Macro-Localization Network and a Micro-Regression Network along with three stacked hourglass as a bottleneck. The proposed model has been tested on three databases $viz.$ FVC2002 DB1\_A, FVC2002 DB2\_A and FPL30K and has been found to achieve true detection rate of 98.75\%, 97.5\% and 92.72\% respectively, which is better than any other state-of-the-art technique.
\end{abstract}

\section{Introduction}
\label{sec:intro}
   Authentication is the need of any secure system. Over the past, we have seen numerous attempts for its automation. Starting from a token-based system, it has grown to include PIN, password and their combinations to recognize a user. Despite all these approaches being efficient and accurate, their susceptibility to loss and leakage makes them fragile. A higher level of security mechanism for the authentication can be achieved by using physiological or behavioral characteristic of the user called biometrics. It uses more intuitive cues such as a face, iris or fingerprint for recognition. Use of these characteristics provides additional advantages in terms of convenience and non-repudiation. Over the period, it has been observed that many biometric traits such as palmprint, knuckle, ear, vein pattern, gait, handwriting, $etc.$ are also suitable for the same \cite{jain2007handbook}.
  
   Fingerprint is an impression that gets develop on a surface touched by the upper part of a human finger \cite{maio2009handbook}. It is one of the most mature among all biometric traits. Evidences are found in ancient clay models that humans at that time were aware of the fact that the fingerprint of every finger is unique. An automatic fingerprint recognition system broadly involves modules for feature extraction, comparison, and decision making. When two fingerprints are given to determine whether they belong to the same person or not, the first step is to extract features from both the fingerprints. Minutiae are one of the standard features that correspond to the location of ridge ending and bifurcation. The comparison module would utilize the location and orientation of minutiae to produce a similarity score, which is the basis for the decision module to categorize whether they are of the same finger or not. This one-to-one comparison can efficiently be performed in constant time for verification.

    \begin{figure} \centering
                    \subfigure {\includegraphics[scale=0.11]{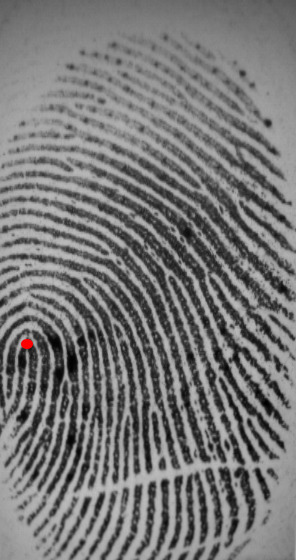}}
                    \subfigure {\includegraphics[scale=0.11]{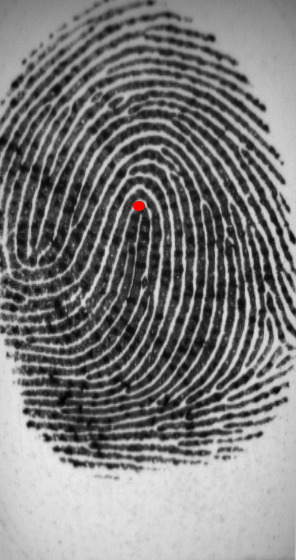}}
                    \subfigure {\includegraphics[scale=0.11]{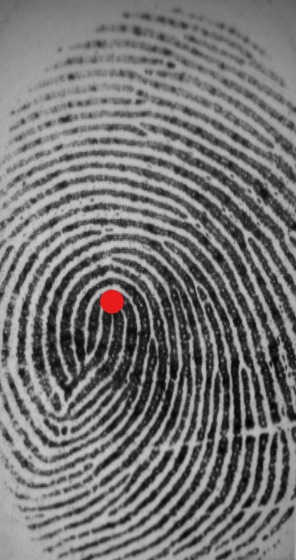}}
                    \subfigure {\includegraphics[scale=0.11]{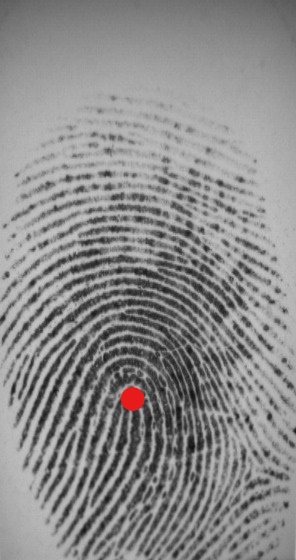}}
                    \subfigure {\includegraphics[scale=0.16]{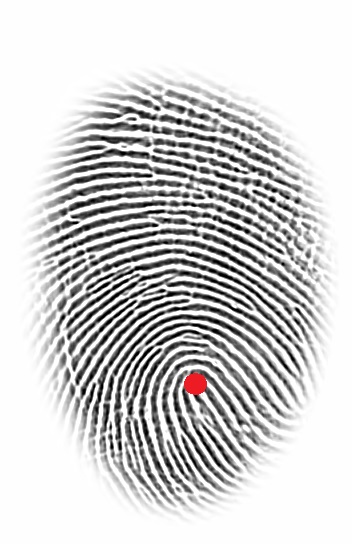}}
                    \subfigure {\includegraphics[scale=0.16]{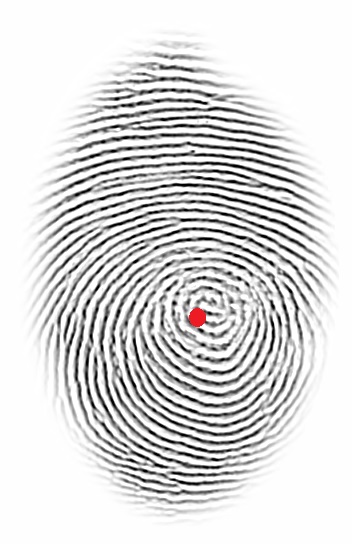}}
                    \subfigure {\includegraphics[scale=0.22]{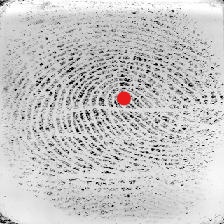}}
                    \subfigure {\includegraphics[scale=0.22]{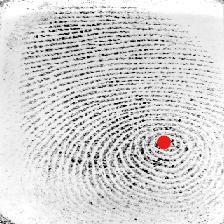}}
                    \hfill
                  \caption{Singular point in a fingerprint image (two each of FVC2002 DB1\_A, \& FVC2002 DB2\_A database and four images from FPL30K database respectively). \label{fig:SampleCorePoints}}
    \end{figure}

   Identification is another interesting application where a single fingerprint is provided to the system, and the target is to determine the identity of the right person in the database to whom this fingerprint should belong. Identification can be performed by applying verification on all the fingerprints in the database. This process is very compute intensive as it involves number of verification proportional to the size of database \cite{tiwari2015indexing}. As the size of the database grows, identification needs more number of comparisons leading to the worsening of system efficiency. To fix the efficiency deterioration, there is a need to develop a strategy that can perform pre-filtering on the database in constant time to produce a small fixed-length candidate set of fingerprints having probabilistic guarantees of hit rate. Such methods are known as Indexing \cite{gupta2018fingerprint}, and they need singular points for normalizing a fingerprint for reliable feature extraction. Singular points are special locations in a fingerprint having high curvature properties. Singular point is shown in some of the fingerprint images of FVC2002 DB1\_A, FVC2002 DB2\_A and FPL30K database in \figurename~\ref{fig:SampleCorePoints}. True and efficient extraction of the singular point plays a major role in successful fingerprint recognition. 

   This paper proposes a novel deep learning based architecture for one shot (end-to-end) singular point detection for a fingerprint image. The model takes a fingerprint image as an input and outputs the location coordinates of the singular point in it. Orientation map could then be used to determine its direction. Proposed model consists of a Macro-Localization Network and a Micro-Regression Network. The Macro-Localization network has a three stacked hourglass as a bottleneck that effectively transforms features. The proposed model has been evaluated on three databases $viz.$ FVC2002 DB1\_A, FVC2002 DB2\_A and FPL30K. Next section reviews important work in singular point detection and highlights novelty of this work.

  \subsection{Related Work}  
   A multi-scale detection algorithm for detecting singular points in a fingerprint image has been proposed in \cite{bo2008fingerprint}. The approach has used Poincare index to detect probable blocks where the singular points could be present and then, used those blocks to identify the location of the singular point. The advantage of this approach is its efficiency as there is no need to calculate the value of Poincare index at every pixel. It only detects the singularities in the effective regions. However, it is susceptible to false detection \cite{iwasokun2014fingerprint}. 
   A hybrid approach using orientation field, directional filtering and Poincare index has been proposed in \cite{gupta2015robust}. This approach has been tested on Biostar, FVC2004 DB1 and FVC2004 DB2 database and achieved an accuracy of 97.80\%, 98.12\%, and 96.38\% respectively. This approach was able to detect singular points in low-quality images or where the singular point is occluded. An approach that uses the walking directional field (WDF), established from the orientation field in \cite{zhu2016walking}. This introduces a walking algorithm that walks to the singular point using the defined WDF instead of scanning the whole fingerprint image. The experimentation has been conducted on subset of FVC2000 DB2, FVC2002 DB2, and FVC2004 DB2 databases. From all the three databases, first fingerprint impression of every subject has been taken, making it a total of 330 fingerprints. Out of these, 30 has been used for training and the remaining 300 fingerprints has been used for testing. The proposed approach achieved 59\% accuracy on the subset of the mentioned databases. 
   An extension of the previous approach has been proposed in \cite{guo2018fast}. 
   While walking on the singularity, the location of the detected singular point is further refined using an enhancement method based on the mean-shift concept.
    
   \begin{figure}
        \centering
        \includegraphics[scale=0.3]{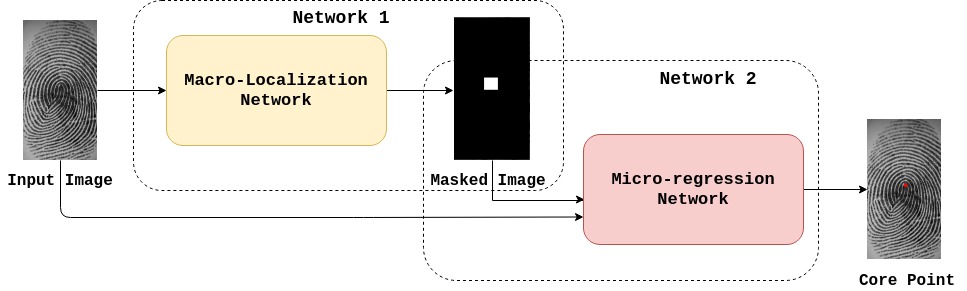}
        \caption{Block Diagram of Proposed SP-Net (dotted lines shows logical parts)}
        \label{fig:pa}
    \end{figure}
    
   An approach utilizing meandering energy potential (MEP) for detection of singular point has been proposed in \cite{tiwari2016meandering}. The advantage of this approach is that it does not require any prior knowledge of the fingerprint structure under evaluation. The proposal has been tested on FVC2002 database. Another approach based on discrete curvature and bending energy that operates in one dimension has been proposed in \cite{zacharias2017fingerprint}. The paper proposes a modified $k$-curvature method to find the high-curvature area of fingerprint ridges. The localization of the reference point is done based on the property of the ridge bending energy. The method has been tested on FVC2002 and FVC2004 datasets. 
   
   A technique to detect singular point utilizing a sliding window by viewing local region of the image has been proposed in \cite{min2018fingerprint}. A window of fixed size, slides through the fingerprint image, to locate the singular point in a partially extracted fingerprint image. The paper uses a complex filter core location method \cite{liang2007algorithm} instead of Poincare Index. The approach on FVC2002 database take 0.2 seconds lesser than a procedure that uses the whole fingerprint image for singular point detection. A singular point detection approach based on Faster-RCNN has been proposed in \cite{liu2018method}. Faster-RCNN \cite{ren2015faster} generates region proposals that potentially contain a singular point. The approach then considered top-$100$ region proposals for singular point detection.
   
    \subsection{Our Contribution} 
    
    To improve the performance of singular point detection, we have developed a novel framework which is a two-step trained end-to-end CNN model, as shown in \figurename~\ref{fig:pa}. This framework consists of two merged networks $viz.$; Macro-Localization Network (MLN) and Micro-Regression Network (MRN). Additionally, MLN is an encoder-decoder network along with three stacked hourglass as a bottleneck. The major contributions of the proposed work are listed as follows:
    
    \begin{itemize}
        \item A novel CNN architecture for singular point detection has been devised from scratch, that has not used any pre-trained weights as in \cite{ipsegnet}.
        
        
        \item Scale-down and scale-up operations have been repeatedly introduced in the stacked hourglass network thus, making the network robust to scale difference, blurriness $etc$.
        
        \item Apart from conducting experiments on standard databases, the proposed model has been tested on an in-house dataset to validate the proposed model.
        
        \item The ground truth for the in-house dataset (FPL30K), $i.e.$ coordinates of the singular point, has been generated manually.
        
    \end{itemize}

   The next section explains the proposed model in detail highlighting the two sub components; macro localization and micro regression network. Subsequent section explains the experimental setting and the obtained results. Conclusions are presented at the end.  
           
    \begin{figure*}
        \centering
        \includegraphics[width=\linewidth,scale=0.35]{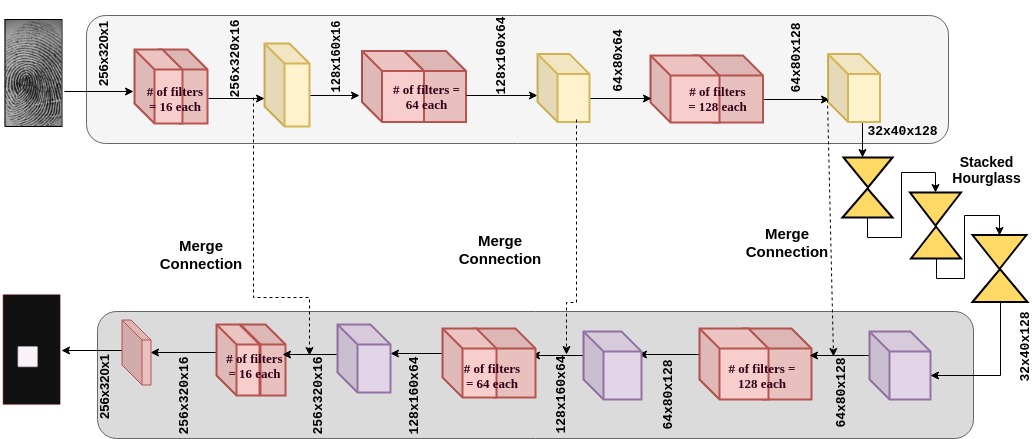}
        \caption{Architecture of Proposed Macro Localization Network }
        \label{fig:segnet}
    \end{figure*}
    
\section{Proposed Approach} \label{sec:pa}

    This section explains the methodology adopted to localize the singular point in a given fingerprint image. There are two major components in this model that are stacked one over the other. The first component applies segmentation and finds the probable region of singular point while the second one applies regression to localize the singular point in the segmented area. The advantage of the proposed model is that it determines the singular point efficiently and in one go. A block diagram depicting the proposed model is given in \figurename~\ref{fig:pa}.

    \subsection{Network Architecture}
    We propose an end-to-end network which takes a fingerprint image as an input and outputs the location of the singular point. It has two sub-modules $viz.$ macro localization and micro regression network as below.
    
    \begin{figure*}
        \centering
         \includegraphics[scale=0.25,angle=90]{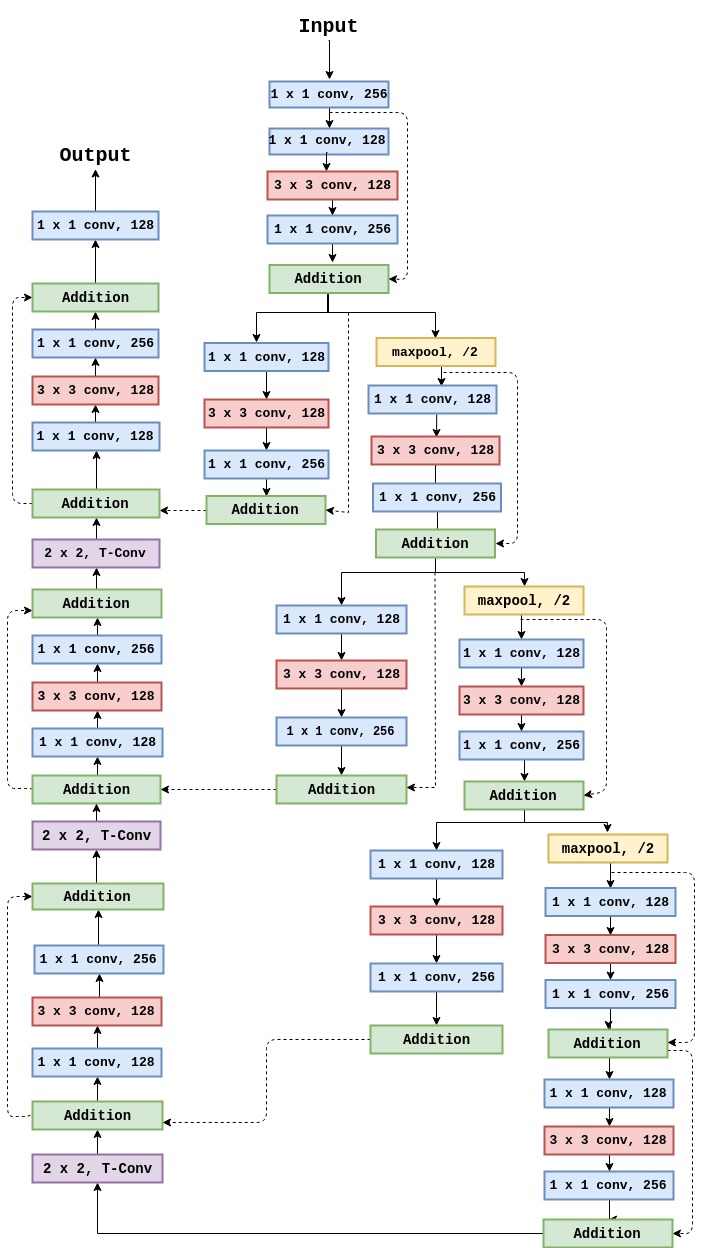}
        \caption{Hourglass Network Used in Proposed Model (Dotted lines represents skip connections)}
        \label{fig:hg}                        
    \end{figure*}

    \noindent\textbf{Macro Localization Network (MLN)}.
      It is an encoder-decoder network that maps the input fingerprint image to an image containing a highlighted region having the highest probability of containing the singular point. The encoder and decoder are based on the U-Net architecture \cite{unet}. The encoder consists of $3${}$\times${}$3$ convolutional layer of filter size 16, 64 and 128. Every pair of convolutional layer is followed by a $2${}$\times${}$2$ max-pool layer. The encoder serves as a feature extractor. The bottleneck is introduced in the form of three-stacked hourglass network between the encoder and decoder. An hourglass network \cite{newell2016stacked} has the ability to capture features at multiple scale and combining them to make pixel-wise predictions. This is made possible by the use of skip-connections that conserve information at every scale. An hourglass network consists of a series of convolution layers followed by a max-pool. This is repeated to process features till a very low scale. The lowest resolution features are then up-sampled. Skip connections are introduced to merge the feature maps at two different scales. The architecture of the single hourglass network is separately shown in \figurename~\ref{fig:hg}. Adding hourglass modules subsequently one after the other allows for re-assessment of the features across the whole image. It also allows for going to-and-fro between the scales which further helps in conserving spatial relationship among features. Directly passing the encoded image to the decoder may result in an inappropriate localization, especially in the noisy images. Adding a stacked hourglass network makes our network robust to such situations. MLN was first tested with adding only one hourglass network as a bottleneck. The network showed improvement in result when two and then three hourglass networks were used. however, there was not much improvement when four hourglass networks were stacked in the bottleneck. Hence, MLN has three stacked hourglass networks as the bottleneck. The output of the bottleneck is passed to the decoder that performs segmentation on the input image. The decoder first performs up-sampling using a $3${}$\times${}$3$ transposed convolution layer. Its output is concatenated with the corresponding feature map of the encoder using merge connections. The merge connections preserve the essential spatial information of the input image that may have lost in the encoder. Using merge connections also eliminates the problem of vanishing gradient during learning. The concatenated feature map is passed to a pair of $3${}$\times${}$3$ convolution layer followed by ReLU activation. This is repeated for different filters of size 128, 64 and 16.     The complete architecture of the network is shown in \figurename~\ref{fig:segnet}. Here red, yellow and violet blocks represent $3${}$\times${}$3$ convolution, $2${}$\times${}$2$ max-pooling, and $3${}$\times${}$3$ up-sampling blocks respectively.

\begin{figure*}
        \centering
        \includegraphics[width=\linewidth, scale = 0.5]{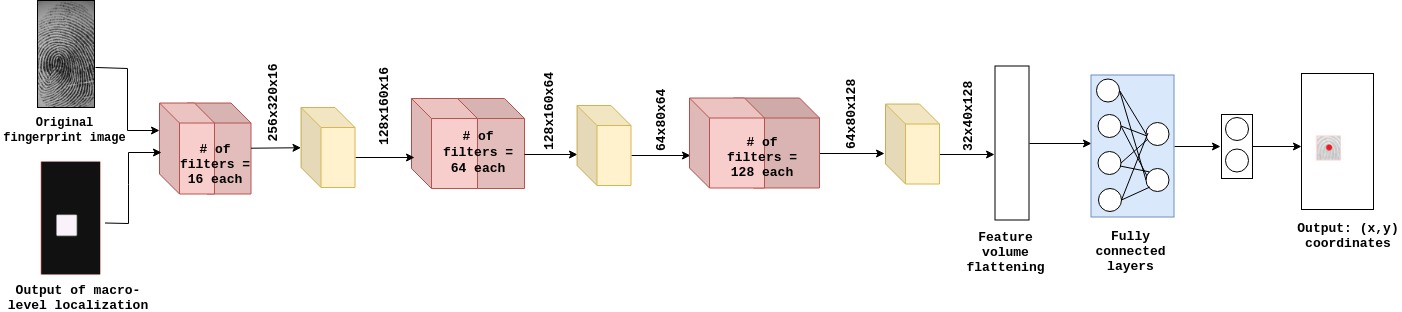}
        \caption{Architecture of  Proposed Micro Regression Network }
        \label{fig:micro-regression}   
    \end{figure*}

    \noindent\textbf{Micro Regression Network (MRN)}.
      This network takes the original fingerprint image along with the output of the previous network $i.e.$ image containing a probable region for singular point proposal as input. The output of MLN gives the probable region that contains the singular point. But, to further trace the location in that region, the original fingerprint image is also passed to MRN that consists of three convolutional blocks with varying filter size 16, 64 and 128 wherein, every block is followed by ReLU activation function and a $2${}$\times${}$2$ max-pool layer. This is succeeded by a flattening and four fully connected layers. The last fully connected layer outputs two values representing the predicted $(x,y)$ coordinates of the singular point. This network mainly performs regression on the proposed region of interest and outputs the location of the singular point in the fingerprint image. The architecture of the Micro Regression network is graphically shown in \figurename~\ref{fig:micro-regression}.
    
      All these component networks are stacked one over the other after training them individually to build a single network that predicts the location of the singular point in a given fingerprint image. Binary cross entropy and mean squared error are used to train the proposed network. The cross entropy loss is back-propagated by the macro-localization network to learn the proposed region for singular point presence while the mean squared error is back-propagated to the regression network for learning singular point localization.   
    
\subsection{Network Training}  
     We have trained our network on three datasets $viz.$ FVC2002 DB1\_A, FVC2002 DB2\_A, and FPL30K. The details of the datasets are given in Section~\ref{sec:database}.Moreover, each dataset has been divided into $80$\%-$20$\% split for training and testing respectively. Training of this network is broadly separated into two parts. First, the auto-encoder with stacked hourglass has been trained. This network takes a fingerprint image as input and produces a mask for fingerprint. During the training of this network, binary cross-entropy loss function has been used. This loss function investigates each pixel of ground truth mask and corresponding predicted mask individually to calculate the loss. Finally, an average loss is computed over all the pixels of the image. For a given fingerprint image $Y$ of size $h${}$\times${}$w$, corresponding predicted mask and ground truth mask are $Y^{pred_{mask}}$ and $Y^{gt_{mask}}$ respectively. The loss between $Y^{pred_{mask}}$ and $Y^{gt_{mask}}$ is defined in Equation~\ref{eq:entropy01}. as:
     
    \begin{eqnarray} \label{eq:entropy01}
         \lefteqn{L_{cross\_entropy}(Y^{pred_{mask}},Y^{gt_{mask}}) = }   \nonumber\\
           &\hspace{80pt} &  \frac{-1}{h*w} \sum_{j=1}^{h} \sum_{i=1}^{w} \Bigg[ Y^{gt_{mask}}(i,j) * \log(Y^{pred_{mask}}(i,j)) + \nonumber\\
           &&\hspace{50pt} (1 - Y^{gt_{mask}}(i,j)) * \log(1 - Y^{pred_{mask}}(i,j)) \Bigg]
    \end{eqnarray} 
 
   In the second step, the regressor network is separately trained by using mean squared error (MSE) as the loss function. We first compute the distance between the actual coordinates and the one predicted for every image. Then total MSE is computed by adding the error for each image and dividing it by the total number of images, as given in Equation~\ref{eq:MSE}. It takes a fingerprint image and corresponding mask concatenated along channel dimension as the input and generates two values depicting the coordinates of the singular point. Later, both the aforementioned networks have been combined by stacking, to form a single end-to-end singular point detection network. 
  
    \begin{equation}
    \label{eq:MSE}
        L_{mean\_squared\_error}(Y^{pred_{coordinates}},Y^{gt_{coordinates}}) =\sum_{i=1}^{n}\frac{{(Y^{pred_{coordinates}}-Y^{gt_{coordinates}}})^2}{n} 
    \end{equation}
  
  \section{Experimental Analysis and Discussions} \label{sec:results}
      This section gives details of the databases used, experimental setting and observations obtained after the experiment. It also specifies the methodology adopted for ground truth generation and evaluation criteria.

     \begin{table}[t]
          \caption{Comparison of the results on FVC2002 DB2\_A with state-of-the-art approaches \label{tab:ComparisonWithOtherTechniques}} \centering 
         \begin{tabular}{p{2.5cm}p{6.5cm}c}\hline
           \cellcolor{intnull2}Technique   & \cellcolor{intnull2}Description &\cellcolor{intnull2}TDR \\\hline\hline
           Zhou et al. \cite{zhou2009novel} & Orientation values along a circle (DORIC) &95.95\% \\\hline
           Xie et al. \cite{xie2010fingerprint} & Inconsistency feature &90.00\%  \\\hline
           Tiwari et al.\cite{tiwari2016meandering} & Meandering energy potential (MEP) & 95.75\% \\\hline
           Liu et al. \cite{liu2018method} & Faster R-CNN &96.03\% \\\hline
            \hline
            \textbf{Proposed} & SP-NET (at less than 10 pixels)   &96.25\% \\\hline
            \end{tabular}\\[8pt]
      \end{table}

    \subsection{Database} \label{sec:database}
      Three databases have been used to evaluate the proposed model. Two of them are publicly available databases that are popularly used for singular point localization. The other one has been collected in-house. The ground truth for all these databases has been generated by manually. The mask of singular point was generated by making a square of size $43\times43$ pixels around the marked ground truth. Description of each of the databases is provided below. 
      
     \noindent\textbf{FVC2002 DB1\_A}. It is a public database \cite{fvc2002Db} widely used for fingerprint recognition benchmark. The database was originally used to host fingerprint verification competition. It contains 800 fingerprint images collected from 100 subjects. Each subject has provided eight fingerprint impressions of the same finger. The databases have been collected using an optical sensors namely, TouchView II by Identix. There exists huge intra-class variation among different fingerprint impressions of the same subject due to a difference in placement, pressure $etc.$
     
     \noindent\textbf{FVC2002 DB2\_A}. This is also a public databases \cite{fvc2002Db} collected in same setting of FVC2002 DB1\_A. It also contains 800 fingerprint images collected from 100 subjects. Each subject has provided eight fingerprint impressions of the same finger. DB2\_A databases have been collected from a different optical sensors namely FX2000 by Biometrika. It also has huge intra-class variation among different impressions of the same subject due to factors such as placement, pressure $etc.$
      
     \noindent\textbf{FPL30K}. This is an in-house database that has been collected from 855 subjects using three different sensors. Subjects are from rural population, involved in laborious field work. Each subject has provided nearly three fingerprint impressions of two of their fingers on each scanner in two phase having gap of two months. Ground truth has been manually generated for all 30,000 fingerprint images. Some of the fingerprint images are shown in \figurename~\ref{fig:SampleCorePoints}. 

    \subsection{Evaluation Parameter} \label{sec:ep}
    True Detection Rate (TDR) has been taken as a measurement index to gauge the performance of the proposed model. The proposed model outputs $(x,y)$ coordinates of the singular point corresponding to the fingerprint sample. The extracted singular point will be considered as a true singular point if the euclidean distance between the original and predicted coordinate is less than or equal to 20 pixels. This can be formulated as given in Equation~\ref{eq:1}, where ($(C_p)^x$,$(C_p)^y$) and ($(C_a)^x$,$(C_a)^y$) refer to the (x,y) coordinates of the predicted and ground truth singular point respectively. 
    
    \begin{equation}
         \sqrt{[(C_p)^x - (C_a)^x]^2+[(C_p)^y - (C_a)^y]^2} \leq 20\;pixels
        \label{eq:1}
    \end{equation}

     \subsection{Experimental Setup}
       The implementation of the proposed network has been done on a Linux based operating system along with NVIDIA GeForce GTX 1080 Ti graphics card with graphics memory of 11 GB. The model has used the Adam optimizer with a learning rate of $0.0005$. The network has been trained for 100 epochs with a mini-batch of 8 images (each of size $256\times 320$).
       
     \begin{figure*}[t]
        \centering
        \includegraphics[scale=0.4]{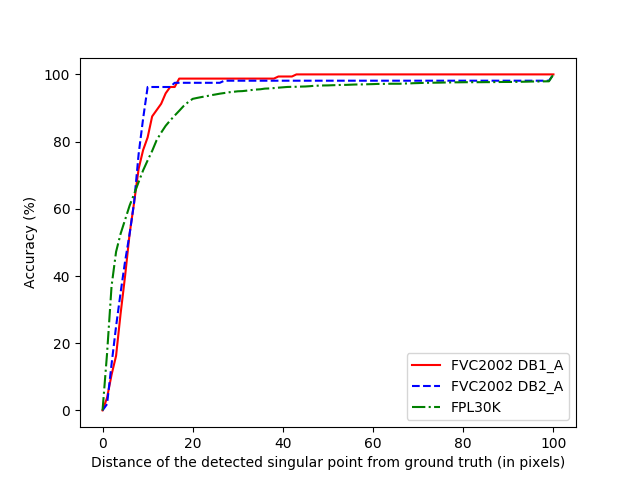}
        \caption{Distance vs Accuracy plots on FPL30K, FVC2002 DB1\_A and FVC2002 DB2\_A}
        \label{fig:graph}                        
    \end{figure*}
       
    \subsection{Observations} \label{sec:ob} 
        
      True detection rate (TDR) for all three databases $viz.$  FVC2002 DB1\_A, FVC2002 DB2\_A, and FPL30K is found to be 98.75\%, 97.5\% and 92.72\% respectively. Results are shown in \tablename~\ref{tab:res_obtained2}. A distance $vs.$ accuracy graph has been plotted for all the databases, and is shown in \figurename~\ref{fig:graph}. The graph shows the effect of increasing tolerance of difference in distance between the ground truth and the predicted coordinates on the model accuracy. 
      
      The proposed model shows a detection rate of 96.25\% on FVC2002 DB1\_A database when the difference in distance between the ground truth and the predicted coordinates is not more then 10 pixels. This has been compared with other state-of-the-art methods and the same is presented in \tablename~\ref{tab:ComparisonWithOtherTechniques}. It can be seen that out model outperforms all of them even when the TDR value was 10 pixels. 
            
\section{Conclusion} \label{sec:conclusion} 
    This paper presents a novel end-to-end deep network to detect singular point for a fingerprint image. The proposed network consists of two logical parts $viz.$ macro-localization and micro-regression. The approach provides an end-to-end network model ensuring efficient computation of singular-point in one go. Experimental results have been obtained on three databases, FVC2002 DB1\_a, FVC2002 DB2\_a and FPL30K containing 800, 800 and 30,000 images respectively. The proposed model was tested on 80\% of these databases individually and tested on the remaining 20\%. The model achieved a TDR of 98.75\%, 97.5\% and 92.72\%, which outperforms any other state-of-the-art technique while still being fast.
    
        \begin{table}[t]
          \caption{Obtained True Detection Rate on Different Databases (20 pixel distance) \label{tab:res_obtained2}} \centering
             \begin{tabular}{c p{3cm} p{2cm} c r}\hline
                   \cellcolor{intnull2}S.No. & \cellcolor{intnull2}Database
                   & \cellcolor{intnull2}Image Count
                   & \cellcolor{intnull2} Trai/Test Partition
                   &\cellcolor{intnull2}TDR \\\hline\hline
                    1. & \textbf{FVC2002 DB1\_A} & 800 & 80-20\% & 98.75\% \\\hline
                    2. & \textbf{FVC2002 DB2\_A} & 800 & 80-20\% & 97.50\% \\\hline
                    3. & \textbf{FPL30K} & 30,000 & 80-20\% & 92.72\% \\\hline
            \end{tabular}
        \end{table}

{\small  \bibliography{paper} }

\begin{thebibliography}{21}
\providecommand{\natexlab}[1]{#1}
\providecommand{\url}[1]{\texttt{#1}}
\expandafter\ifx\csname urlstyle\endcsname\relax
  \providecommand{\doi}[1]{doi: #1}\else
  \providecommand{\doi}{doi: \begingroup \urlstyle{rm}\Url}\fi

\bibitem[fvc()]{fvc2002Db}
The {FVC2002} database:.
\newblock http://bias.csr.unibo.it/fvc2002/.

\bibitem[Bo et~al.(2008)Bo, Ping, and Lan]{bo2008fingerprint}
Jin Bo, Tang~Hua Ping, and Xu~Ming Lan.
\newblock Fingerprint singular point detection algorithm by poincar{\'e} index.
\newblock \emph{WSEAS Transactions on Systems}, 7\penalty0 (12):\penalty0
  1453--1462, 2008.

\bibitem[Guo et~al.(2018)Guo, Zhu, and Yin]{guo2018fast}
Xifeng Guo, En~Zhu, and Jianping Yin.
\newblock A fast and accurate method for detecting fingerprint reference point.
\newblock \emph{Neural Computing and Applications}, 29\penalty0 (1):\penalty0
  21--31, 2018.

\bibitem[Gupta et~al.(2019)Gupta, Tiwari, and Arora]{gupta2018fingerprint}
Phalguni Gupta, Kamlesh Tiwari, and Geetika Arora.
\newblock Fingerprint indexing schemes - a survey.
\newblock \emph{Neurocomputing}, 335:\penalty0 352--365, 2019.

\bibitem[Gupta and Gupta(2015)]{gupta2015robust}
Puneet Gupta and Phalguni Gupta.
\newblock A robust singular point detection algorithm.
\newblock \emph{Applied Soft Computing}, 29:\penalty0 411--423, 2015.

\bibitem[Iwasokun and Akinyokun(2014)]{iwasokun2014fingerprint}
Gabriel~Babatunde Iwasokun and Oluwole~Charles Akinyokun.
\newblock Fingerprint singular point detection based on modified poincare index
  method.
\newblock \emph{International Journal of Signal Processing, Image Processing
  and Pattern Recognition}, 7\penalty0 (5):\penalty0 259--272, 2014.

\bibitem[Jain et~al.(2007)Jain, Flynn, and Ross]{jain2007handbook}
Anil~K Jain, Patrick Flynn, and Arun~A Ross.
\newblock \emph{Handbook of biometrics}.
\newblock Springer Science \& Business Media, 2007.

\bibitem[Liang et~al.(2007)Liang, Zhao, He, and Tian]{liang2007algorithm}
L~Liang, H~Zhao, P~He, and H~Tian.
\newblock Algorithm of extracting core point in fingerprint.
\newblock \emph{JOURNAL-HEBEI UNIVERSITY OF TECHNOLOGY}, 36\penalty0
  (1):\penalty0 46, 2007.

\bibitem[Liu et~al.(2018)Liu, Zhou, Han, Guo, and Qin]{liu2018method}
Yonghong Liu, Baicun Zhou, Congying Han, Tiande Guo, and Jin Qin.
\newblock A method for singular points detection based on faster-rcnn.
\newblock \emph{Applied Sciences}, 8\penalty0 (10):\penalty0 1853, 2018.

\bibitem[Maio and Jain(2009)]{maio2009handbook}
Dario Maio and Anil~K Jain.
\newblock \emph{Handbook of fingerprint recognition}.
\newblock springer, 2009.

\bibitem[Min et~al.(2018)Min, Zhang, and Ren]{min2018fingerprint}
Xiangshen Min, Xuefeng Zhang, and Fang Ren.
\newblock Fingerprint core location algorithm based on sliding window.
\newblock \emph{Wuhan University Journal of Natural Sciences}, 23\penalty0
  (3):\penalty0 195--200, 2018.

\bibitem[Newell et~al.(2016)Newell, Yang, and Deng]{newell2016stacked}
Alejandro Newell, Kaiyu Yang, and Jia Deng.
\newblock Stacked hourglass networks for human pose estimation.
\newblock In \emph{European Conference on Computer Vision}, pages 483--499.
  Springer, 2016.

\bibitem[Patil et~al.(2017)Patil, Jha, and Nigam]{ipsegnet}
Shreyas~Malakarjun Patil, Ranjeet~Ranjan Jha, and Aditya Nigam.
\newblock Ipsegnet: Deep convolutional neural network based segmentation
  framework for iris and pupil.
\newblock In \emph{International Conference on Signal-Image Technology \&
  Internet-Based Systems (SITIS)}, pages 184--191. IEEE, 2017.

\bibitem[Ren et~al.(2015)Ren, He, Girshick, and Sun]{ren2015faster}
Shaoqing Ren, Kaiming He, Ross Girshick, and Jian Sun.
\newblock Faster r-cnn: Towards real-time object detection with region proposal
  networks.
\newblock In \emph{Advances in neural information processing systems}, pages
  91--99, 2015.

\bibitem[Ronneberger et~al.(2015)Ronneberger, Fischer, and Brox]{unet}
Olaf Ronneberger, Philipp Fischer, and Thomas Brox.
\newblock {U-Net}: Convolutional networks for biomedical image segmentation.
\newblock In \emph{International Conference on Medical image computing and
  computer-assisted intervention}, pages 234--241. Springer, 2015.

\bibitem[Tiwari and Gupta(2015)]{tiwari2015indexing}
Kamlesh Tiwari and Phalguni Gupta.
\newblock Indexing fingerprint database with minutiae based coaxial gaussian
  track code and quantized lookup table.
\newblock In \emph{International Conference on Image Processing}, pages
  4773--4777. IEEE, 2015.

\bibitem[Tiwari and Gupta(2016)]{tiwari2016meandering}
Kamlesh Tiwari and Phalguni Gupta.
\newblock Meandering energy potential to locate singular point of fingerprint.
\newblock In \emph{International Conference on Biometrics (ICB)}, pages 1--6.
  IEEE, 2016.

\bibitem[Xie et~al.(2010)Xie, Yoo, Park, and Yoon]{xie2010fingerprint}
SJ~Xie, HM~Yoo, DS~Park, and S~Yoon.
\newblock Fingerprint reference point determination based on a novel ridgeline
  feature.
\newblock In \emph{Intl. Conference on Image Processing}, pages 3073--3076.
  IEEE, 2010.

\bibitem[Zacharias et~al.(2017)Zacharias, Nair, and
  Lal]{zacharias2017fingerprint}
Geevar~C Zacharias, Madhu~S Nair, and P~Sojan Lal.
\newblock Fingerprint reference point identification based on chain encoded
  discrete curvature and bending energy.
\newblock \emph{Pattern Analysis and Applications}, 20\penalty0 (1):\penalty0
  253--267, 2017.

\bibitem[Zhou et~al.(2009)Zhou, Chen, and Gu]{zhou2009novel}
Jie Zhou, Fanglin Chen, and Jinwei Gu.
\newblock A novel algorithm for detecting singular points from fingerprint
  images.
\newblock \emph{IEEE Transactions on Pattern Analysis and Machine
  Intelligence}, 31\penalty0 (7):\penalty0 1239--1250, 2009.

\bibitem[Zhu et~al.(2016)Zhu, Guo, and Yin]{zhu2016walking}
En~Zhu, Xifeng Guo, and Jianping Yin.
\newblock Walking to singular points of fingerprints.
\newblock \emph{Pattern Recognition}, 56:\penalty0 116--128, 2016.

\end{thebibliography}

\end{document}